# Clustering and Relational Ambiguity: From Text Data to Natural Data

Nicolas Turenne*

INRA, SenS, UR1326, Université Paris Est, Champs-sur-Marne, F-77420, France

*Corresponding author: Nicolas Turenne nturenne_inra@yahoo.fr

**Abstract**
Text data is often seen as "take-away" materials with little noise and easy to process information. Main questions are how to get data and transform them into a good document format. But data can be sensitive to noise often called ambiguities. Ambiguities are aware from a long time, mainly because polysemy is obvious in language and context is required to remove uncertainty. One claim in this paper is that syntactic context is not sufficient to improve interpretation. This paper tries to explain that firstly noise can come from natural data themselves, even involving high technology, secondly texts, seen as verified but meaningless, can spoil content of a corpus; it may lead to contradictions and background noise. We used a set of papers in biology to identify ambiguous facts related to human interpretation and a nearest-neighbour display associated to a Zipfian distribution to compare structural content of a corpus. Four kinds of discourse technical, general, short-communication and artificial have been studied.

## INTRODUCTION

Human cognition refers to a diversity of concepts such as memory and brain anatomy, inference and reasoning, motivation, time and space, classification and clustering. Inference tries to identify good relations or properties associated to an object. In this sense it is also possible to test validity or consistency of a relation. Let be the proposition P = "a cat is a stone" is false or contradictory because a stone is not a living organism, though a cat is a living organism. P can be called paradoxal or contradictory. Sometimes, society lives with contradictions such as tolerance to many deaths on roads or in wars but intolerance for death from diseases. In this paper we more specifically focus on sources of potential contradictions that could spoil computation of information extraction.

Formal semantics is attached to validate relations between a set of objects. Our study focuses on issues in managing complexity of a logical proposition and how to compute its truth value, but we also study how to extract relations and see how they are asserted as non contradictory with regard to other relations extracted in other texts. Thus texts are the primary material of discussion.

Chapter 1 presents relational ambiguities we can find in text. We start by presenting a typology of logical relations. Given a type of relations, we explain how to extract such relations with markers in text files. But markers are not sufficient to detect a contradiction. A specialized language such as molecular corpus provides an example of ambiguous relations (contradictory) that cannot be detected with markers. Hence, we show that a global overview of words collocations in a corpuscan give a good signal about the structure. In our "publish or perish" new era of research and development system, production of literature is high but a non-negligible percent of papers becomes false over time. It is possible to compile from the PubMed website 4,800 papers honestly accepted but hence officially retracted. Such amount of information is original to make a corpus of real texts, written with the intention to propose arguments and content to readers, in a real natural language, but knowing that the content has been invalidated ex-post by readers. To build a random text is quite easy but the grammar and arguments will not be normalized by how argumentation is made usually by "normal writers and experts", or natural language will not confirm standard use of official grammar and texts normally used to make corpora. In this sense we can consider such texts as "well-constructed" in the sense we can find such texts in nature (on official databases) and purely noisy.

Chapter 2 presents a source of ambiguity coming from human interpretation of natural data. Scientific and technological texts are supposed to take their foundation from validated experimental devices producing experimental data. A device can be technological as a telescope in astronomy or a survey form in psychology. We present an overview of ambiguous interpretation across several sciences, which should impact conclusions of practicians and the way they can restitute results in documents. We cannot call it birth of controversy but ambiguous interpretability of data of specific results requiring validation by other techniques. Often a controversy occurs when several techniques lead to opposite conclusions, as a protocol is supposed to be scientific when it gives a warranty of results reproductibility.

# I RELATIONAL AMBIGUITY IN DATA

## 1.1 Antinomy and paradox in mathematics

In philosophy and logics, paradox has been attributed to Greek rhetoric during the VII century before JC. The first paradox has been the liar of Epimenide. It says that "a man told that he was lying. What he said was true or false?" In another way let us reformulate in this way, Epimenide says "All Cretan are liars". This was considered by antic philosophers as a paradox. Anyway, Epimenide tells the truth, then he lies (because he is Cretan), so the statement is false (because all Cretans lie). Otherwise, in the contrary, if Epimenide lies by saying that, then its statement is false: there is at least one Cretan telling the truth, what is not contradictory, because it is the solution of the paradox.

In modern mathematics, the logician Russel described the following paradox in 1902 formulated by this question: "is the class of all classes which are not element of themselves, element of itself?" In 1919 he reformulated the statement in a vernacular language such as "The Barber of a given village shave exactly each person who does not shave himself. Question: does this barber shave himself?"

If we search for a solution in a predicate analysis framework the reasoning leads to a contradiction.

Let R = {x such that x is not an element of x}, If R is an element of R then R matches "x is not element of x", hence R is not an element of R. Thus contradiction.
If R is not element of R then R does not match "x is not element of x", which means R matches non-"x is not element of x" what is equivalent to "x is element of x" so R is an element of R. Contradiction.

The theory of sets permits escaping the contradiction because a set cannot contain itself.

## 1.2 Antinomy in linguistics

Poetry is, and was for long, a playground for using words not usually used in the same context such as in French: "Dans un temps proche et très lointain" or "Je suis et je ne suis plus". More radically in any language we can find pairs of words associating contrary meanings. Most of them are verbs and adjectives such as: to move back / to move forward, to begin / to stop, to increase / to decrease, black / white, elitist / popular, fast / slow, big / small, wet / dry. We can also find what is called quasi-antonyms such as bon/terrible.

According to Antoine Culioli [Culioli, 1987], contrary pairs are an illusion of language, which better tends to construct complementary pairs in the sense of mathematical logics, such as "white" and "non-white" meaning any colour except white. For A. Culioli, fuzzy sets should be an interesting framework but such formalism is too weak for a fine description.

Recall families of linguistic antonomy [Herrmann et al, 1986]. Two lexical items are linked by antonomy relationship if it is possible to draw a symmetry of their semantic features through an axis. Symmetry can be defined in different ways, according to the nature of the support. We observe several supports setting each one in a different antonomy:

- complementary antonomy concerns application (or non-application) of a property ('applicable' / 'non-applicable', 'presence' / 'absence' ): for instance, 'shapeless' is an antonym of all having a form, also 'tasteless', 'colourless', 'odourless', etc. is about all having taste, colour, smell. …In classical logics definition is $(\forall x)[P(x) \Rightarrow \neg Q(x)]$

- scalar antonomy concerns a property influencing a scalable value (high value, low value): for instance, 'hot', 'cold' are symmetrical values of temperature; It is explained by the existence of a "neutral value" from which the others are settled. In classical logics it can be expressed by $(\forall x)[P(x) \Rightarrow \neg Q(x) \wedge R(x)]$ if R is the property having a reference value (neutral or median)

- dual antonomy is concerned by existence of a property or an element considered as symmetrical by usage (for instance 'sun', 'moon', or by natural or physical properties about studied objects (for instance 'male', 'female', 'head', 'foot', …); $(\forall x, y)[P(x,y) \Rightarrow Q(y,x)]$

Usage of textual resources such as corpora occurred in the domain of psychology in 1989 with studies by Charles and Miller [Charles and Miller, 1989] aiming to check, with the help of the Brown Corpus, the hypothesis of Deese: two adjectives with opposite meaning are supposed to be antonyms when they are considered switchable over most of their contexts [Deese, 1965]. About twenty-five years later, [Justeson and Katz, 1991], [Fellbaum, 1995], [Willners, 2001] and [Jones, 2002] have defined a set of morpho-syntactic schemes to detect automatic antonyms candidates (see Table 1). Such patterns can also be defined in other languages as French [Amsili, 2003] (see Table 2).

| (both) X and Y | from X to Y | X gives way to Y |
| X as well as Y | how X or Y | X not Y |
| X and Y alike | more X than Y | X instead of Y |
| neither X nor Y | X is more ADJ than Y | X as opposed to Y |
| (either) X or Y | the difference between X and Y | the very X and the very Y |
| X rather than Y | separating X and Y | either too X or too Y |
| whether X or Y | a gap between X and Y | deeply X and deeply Y |
| now X, now Y | turning X into Y | |

Table 1. Morpho-syntactic schemes to extract antonyms in English.

| X ou Y | de/depuis X à/jusqu'à Y | X comme Y |
| « diurne ou nocturne » | « depuis les racines jusqu'aux feuilles » | « parisiens comme provinciaux » |
| soit X soit Y | ni X ni Y | plus/moins/aussi X que Y |
| « soit constante, soit croissante » | « ni implicitement ni explicitement » | « plus symbolique que réel » |
| à la fois X et Y | aussi bien X que Y | |
| « à la fois offensives et défensives » entre X et Y | « aussi bien physiquement que mentalement » X plutôt que Y | Variation du patron: aussi bien X que Y ► aussi bien défensif qu'offensif |
| « entre exigences et besoins » | « comprendre plutôt que juger » | |

Table 2. Morpho-syntactic schemes to extract antonyms in French.

## 1.3 Ambiguity in a specialized domain

*1.3.1 Relational ambiguity*

Table 2 presents what should look as opposition in texts according to linguistic markers and rhetorical expression. If we focused on a specialized discourse, opposition could take another expression. [Reinitz et al, 1998] have studied, in molecular biology, the fly species and shown an ambiguity in the role of the *SmaI-BglI* protein to create stripe 6 in the fly body.

According to [Howard and Struhl, 1990] "*Further deletion analysis of this region (particularly constructs ET44, 30 and 31) provides clear evidence that an 600 bp **region of DNA (from position -8.4 to -9.0**; ET31) contains all of the elements necessary and sufficient for a relatively normal stripe 6 response (Fig. 3B). However, we note that this response seems to be displaced slightly **posterior to** the location of the endogenous **stripe 6** at this stage*".
But according to [Langeland et al, 1994] "*The **526 bp SmaI-BglI** reporter construct (6(526)lacZ) gives **rise to** strong lacZ stripe expression corresponding to h **stripe 6***".

Another example of contradiction is the one pointed out by [Giles and Wren, 2008] noting behaviour uncertainty between *c-jun* and *c-myc* genes.

According to [Davidson et al, 1993] "*17 bet – Estradiol had little effect on expression of c-jun, jun B, jun D, or c-fos mRNA by MCF-7 cells over 12 h, although it stimulated c-myc expression 4-fold within 30 min*".
But [Bhalla et al, 1993] formulated differently: "*In addition, intracellularly, mitoxantrone-induced PCD was associated with a marked induction of c-jun and significant repression of c-myc and BCL-2 oncogenes*".

*1.3.2 Visual ambiguity*

As a consequence of ambiguous morphological similarities, many species have been moved between genera or even families since the earliest exhaustive classifications of liverworts [Hentschel et al, 2007]. According to [Carette and Ferguson, 1992] both programmed cell death, and in particular epithelial-mesenchymal transformation theory of seam degeneration rely on the potentially ambiguous interpretation of a dynamic event from a series of static images. In phylogenetics, [Yu et al, 2010] pointed out an ambiguous interpretation about inference for the entire cladogram. [Palomares-Ruis et al, 2010] advocate for phylogenetic relationships within plant-parasitic nematodes such as Longidoridae, especially in cases where morphological characters may lead to ambiguous interpretation. [Gantchev et al, 1992] advocate for the spin-labelling technique but recall that in studying the dynamic behaviour of biological membranes an unambiguous interpretation of the spectral data is difficult. [Ivanov, 2004] wrote about echolocation of dolphins by imagery and describes that if the animal changes the spectral-time structure of echolocation pulses on purpose, the statistical processing yields an ambiguous interpretation of data on the acoustic behaviour of a dolphin in the course of the detection and identification of targets. [Shin and Pierce, 2004] warn about difficult interpretation of the fluorescence signal caused by fluorescence resonance energy transfer between dyes. [Da Silva and Oliveira, 2008] criticize the ERIC-PCR technique devoted to identification of strain groups, due to interpretation limitations leading to low reproducibility between laboratories. [Gorbatyuk et al, 1996] point out that 1H nuclear magnetic resonance (NMR) spectra were assigned incorrectly because of a rather ambiguous interpretation of the spectra in absence of the complementary 13C NMR spectra. [Wood and Napel, 1992] discuss radiological imagery interpretation problems about surface orientation of the reconstructed objects though this problem can be avoided by using multiple light sources.

*1.3.3 Measure ambiguity*

The use of multiple molecular markers as aids in genetic selection programs can be spoiled due to collinearity [Gianola et al, 2006]. Some DNA sequences such as 16S rRNA sequencing may occur in species harbouring multiple copies of the 16S rRNA gene, as demonstrated between the different operons in E. coli [Mollet et al, 1997]. The importance of unequivocal annotation of microarray experiments is evident. The different probe and gene IDs corresponding to the two annotation releases generate uncertainties [Noth and Benecke, 2005]. PCR methods can sometimes be controversial and a post-PCR control has been shown to be often essential to confirm a sequence identity in case of ambiguous recognition of specific targets [Peano et al, 2005]. In some biological approaches ionophores were used for demonstration of the electrogenic properties of the enzyme, which could lead to a problem of interpretation of electrogenicity [Eisenrauch and Bamberg, 1990]. [Kloczkowski et al, 2002] recall that hydrogen bond placement can be different because of ambiguous interpretation of imperfect geometries inherent in experimental structures. Diagnosis relies on techniques, one of them is serology. In spite of high sensitivity, routine serological tests provide results of ambiguous interpretation [Kompalic-Cristo et al, 2004]. Occasionally, unwanted nonspecific PCR products, often in the size range of the expected product, are obtained during the amplification process; this can lead to ambiguous interpretation of results in ethidium bromide-stained gel analyses [Battles et al, 1995]. [Moskovets et al, 2003] related a weak fragmentation of singly charged precursors in MALDI TOF/TOF-MS (compared with collision-induced fragmentation of doubly charged precursors in ESI-MS) often provides only a few fragment peaks, resulting in ambiguous interpretation. Typical and conventional methods to detect E. coli are cultivation of the organism in selective media and identification by their morphological, biochemical and immunological characteristics. Because of ambiguous interpretation of the results [Won and Min, 2010] recommend long detection times from initiation to readout, and relatively low detection limits of the cultivating methods using selective media. To study epidermal UV absorption of leaves from chlorophyll fluorescence measurements, [Ounis et al, 2000] explain that fluorescence emission ratios (Blue/Red or Blue/Far-Red) present a limitation because they depend on two variables, which can vary independently, leading to ambiguous interpretation.

In this part, we follow the ideas of the first chapter to draw up where ambiguity and source of confusion is contained in data. Lots of scientific research, producing information and data may induce an expert in a confusing position to offer a precise interpretation. We cited for each scientific discipline a pool of studies that is only representative, but not exhaustive, of occurring problems. For ambiguities in texts the two effects of misinterpretation and erroneous results can play an important role.

## II COMPARISON OF REAL AND ARTIFICIAL CORPORA

**2.1 Data**

We try to compare the lexical distribution and associations between an artificial corpus and a real corpus about the same size.
From [Sinclair, 1991] we pick up a general definition of a corpus.

*Definition 1: Corpus*
A collection of naturally occurring language text, chosen to characterize a state or variety of a language.

Hence we set the following definition:

*Definition 2: Specialized Corpus*
A specialized corpus is written in a human vernacular language. It has to cover all discussions of a technical field from past to present.

If we take definition 2, a specialized corpus covers whatever people can say about the field. A specialized corpus contains all relations of a given field. Suppose two corpora C1 and C2 are specialized of the same field, if a relation is contained in C1 but not in C2, it means that C2 does not cover the field; to make a real corpus of the field, C1 and C2 have to be merged, or C2 can be called a sub-corpus of the field. According to that we cannot compare a specialized corpus with another of the same field. But we can create artificial corpora. An artificial corpus is influenced by lexical composition and the grammar it uses.

Our hypothesis here is that lexical distribution and grammar can lead to a different density of relation. Hence we expect that only a specialized corpus will give a relational structure similar to another specialized corpus only.

We used ten different corpora and one specialized corpus. We made the ten corpora in terms of the size of documents or words of the specialized corpus. Among the ten corpora, four are artificial, they are settled with a mixture model (lexical distribution and grammar):
- "Corpus BD" is a real corpus that is specialized about the biodiversity domain. It contains 4,655 abstracts of projects. Is has been created from the BIODIVERSA database containing 6,500 projects with duplicates [http1, 2013].
- "Corpus PM" is an artificial corpus built from 4,835 PubMed retracted articles (Corpus PM). They form real texts written as proofed natural language but their content is false since they have been retracted from journals where they were published initially. It has been created from the PubMed database containing 21 million publications [http2, 2013].
- "Corpus TC" is a corpus of 6,049 abstracts of patents. We used 121 major codes of the IPC (international patent classification) ontology. For each code we kept 50 patents as a uniform mixture model of topics. It has been created from the EPO website containing 78 million patents [http3, 2013].
- "Corpus SCI" is a corpus of 5,111 abstracts of scientific papers. We used 52 major academic sections by the French Ministry of Research. For each code we kept 100 publications as a uniform mixture model of topics. It has been built with the Web Of Science database containing 45 million publications [http4, 2013].
- "Corpus RD" is a corpus of 6,502 generated abstracts containing at least 150 words from an automatic random text generator in the marketing field. Used tool is called the "corporate bullshit generator" using a dependency grammar of basic sentences and a tree of 781 lexical items [http5, 2013].

- "Corpus CS" is a corpus of 8,000 generated abstracts from an automatic text generator called SCIgen, which can generate artificial papers in computer science with figures and citations. A generated paper has been accepted in 2005 to WMSCI, the World Multiconference on Systemics, Cybernetics and Informatics [http6, 2013].
- "Corpus SL" is a corpus of 6,500 generated abstracts containing 150 words from randomized sequences generated randomly by a list of words. The list of words comes from the Wikipedia [http7, 2012] version from the 28th April 2012. It contains 156,209 words. From this list we selected a subpart of 1,000 words to generate sentences.
- "Corpus BL" is a corpus of 6,500 generated abstracts containing 150 words from randomized sequences generated randomly in the same way as "corpus SL" but with an extended lexical dictionary of about 50,000 words.
- "Corpus NG" is a corpus of 3,971 among the 18,846 from 20 newsgroups from web forum exchanges.
- "Corpus RT" is a corpus of Reuter's news. 6,025 news items were kept among the 21,578 of the collection.
- "Corpus TW" is a corpus consisting of 50,000 tweets in English. The length of each tweet is about 15 words. It comes from the Twitter database.

## 2.2 Word distribution

The distributional study of frequent words leads to the capture of much information through the most significative lexical items under the hypothesis of their high-level of repetitive occurrences. The pioneering work of Georges Zipf shows a typical distribution $x.y = Constant$ where $x$ is the sorted rank over frequency and the y-axis is the number of occurrences of elementary lexical items in a long text or a set of texts in a given language [Zipf, 1935]. Frequency is defined by the number of occurrences of a lexical item in the corpus.

We used the R platform [R Core Team, 2013], and especially the tm package [Feinerer et al, 2008] and basic matrix functions, to split corpora into elementary lexical items and to sort frequent items. Punctuation, figures and words smaller than three characters had been deleted. When stemming the raw text, we kept only the root form of each word and the text is less dense as seen in Table 3.

| *The project envisages to continue and extend the studies of evolution and systematics in the grass genus Bromus and the legume genus Vicia supported by our ending Estonian Science Foundation Grant No.4082 for 2000-2003. The project combines traditional morphology-based botanical systematics, biochemical isozyme analyses, genetics, and chromosome cytology for solving problems of phylogenetic systematics, phylogeography and evolution in botany. The main objectives of the project are: 1. To determine genetic divergence and relationships within and among species Bromus hordeaceus, B. secalinus, B. racemosus and B. commutatus of type section of genus Bromus by cladistic and phenetic analysis of isozymes with checking the correspondence of the results with the traditional morphological species delimitation* | project envisag continu extend studi evolut systemat grass genus bromus legum genus vicia support estonian scienc foundat grant project combin tradit morphologybas botan systemat biochem isozym analys genet chromosom cytolog solv phylogenet systemat phylogeographi evolut botani main object project determin genet diverg relationship speci bromus hordeaceus secalinus racemosus commutatus type section genus bromus cladist phenet analysi isozym check correspond result tradit morphology speci delimit |
|---|---|

Table 3. Part of text from corpus BD in raw form (left) and stemmed form (right).

Table 4 shows statistics about word counts over all the used corpora. There are zero values because some corpora are randomly generated and do not follow the standard law observed in natural corpora. This is also summarized and discussed later in chapter 2.5.

|  | #documents | #words | #words AND freq>1 | #words AND freq=2 | #words AND freq=3 | #lemmatized words | #lemmatized words AND freq>1 | #lemmatized words AND freq=2 | #lemmatized words AND freq=3 |
|---|---|---|---|---|---|---|---|---|---|
| Corpus BD | 4,655 | 36,023 | 19,588 | 4,966 | 2,509 | 25,875 | 13,519 | 3,568 | 1,734 |
| Corpus PM | 4,835 | 29,044 | 16,882 | 4,230 | 2,419 | 22,440 | 12,866 | 3,150 | 1,800 |
| Corpus TC | 6,049 | 21,647 | 14,014 | 3,288 | 1,777 | 13,924 | 8,821 | 2,077 | 1,078 |
| Corpus SCI | 5,111 | 42,248 | 23,808 | 6,697 | 3,495 | 30,106 | 16,696 | 4,770 | 2,481 |
| Corpus RD | 6,502 | 981 | 981 | 0 | 0 | 623 | 623 | 0 | 0 |
| Corpus CS | 8,000 | 1,533 | 1,497 | 33 | 40 | 1,430 | 1,395 | 33 | 39 |
| Corpus SL | 6,000 | 973 | 973 | 0 | 0 | 971 | 971 | 0 | 0 |
| Corpus BL | 6,000 | 20,505 | 20,505 | 0 | 0 | 19,790 | 19,790 | 0 | 0 |
| Corpus RT | 6,025 | 68,007 | 22,254 | 8754 | 3256 | 59,776 | 18,407 | 7,621 | 2,783 |
| Corpus NG | 3,971 | 78,713 | 28,469 | 9,998 | 3,819 | 66,325 | 21,812 | 8,090 | 2,926 |
| Corpus TW | 50,000 | 52,840 | 17,421 | 6,214 | 2,676 | 46,323 | 14,565 | 5,219 | 2,195 |

Table 4. Distribution over corpora of words count, lemmas count, words occurring three times, words occurring two times, and words occurring more than one time. Grey line represents the reference corpus about biodiversity.

Regardless of the kind of corpus, with or without stemming, words occurring one time represent between 41.9 and 54.4% of all features, words occurring two or three times represent between 38.7 and 47.7% of all features occurring more than one time, or between 19.5 and 21.8% of the whole set of items (see Figure 1).

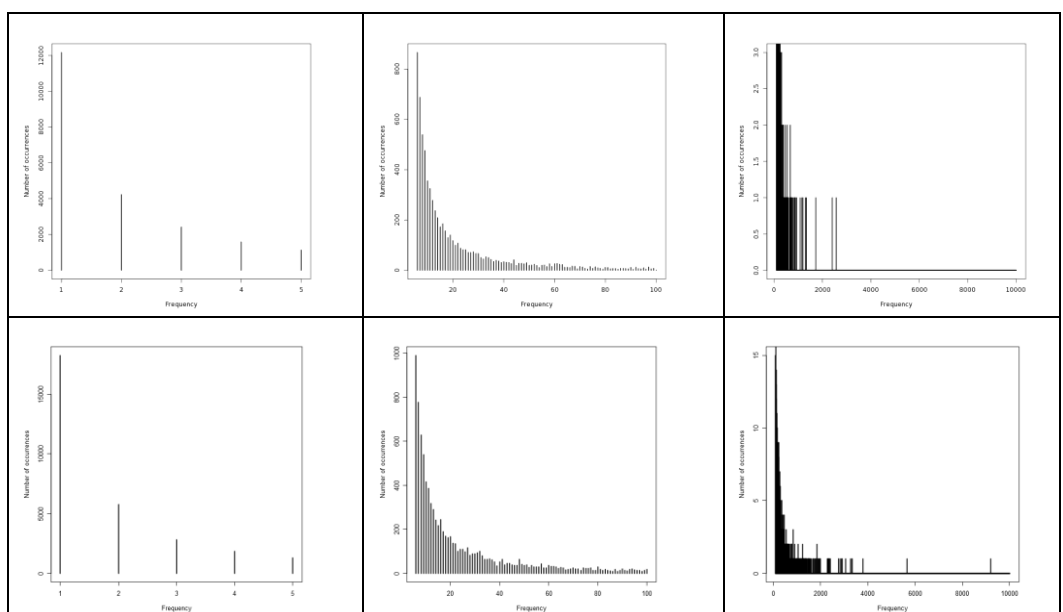

Figure 1. Distribution of number of lexical items (y-axis) and according to the occurrence number (x-axis). Upper line shows the PM corpus: occurrence range 1-5 (upper left), 6-100 (upper middle), 101-2000 (upper right). It contains 12,162 itms with one occurrence. Bottom line shows the BD corpus: occurrence range 1-5 (bottom left), 6-100 (bottom middle), and 101-10000 (bottom right). It contains 18,228 items with one occurrence.

Being aware of the large amount of items and their distribution frequency can offer an anchoring to catch strong lexical semantic signals about the content [Sparck-Jones, 1972]. For relevant feature extraction, a basic process relies on frequent items selection. We set a threshold to make a comparison of the itemset extraction. A reasonable figure should be 5% of documents, but we decided to show the occurrences distribution over three representative thresholds: 2 (minimal value), 100 (medium value) and 1000 (high value). We call this number $S_f$, knowing that an item can occur several times in the same document.

Table 5 gives results from two corpora: the BD corpus about the biodiversity abstracts project and the PM corpus about retracted abstracts from the PubMed database. Amount of frequent words is low with a high $S_f$ value. When words are stemmed the global amount of lexical items decreases but increases with $S_f= 1000$ because of concatenation of some lexical variants. In Table 6 we show the top list of stemmed words for both corpora. This kind of list is quite powerful to get a crude idea of the content of a corpus.

|  | Corpus BD | | | Corpus PM | | |
| --- | --- | --- | --- | --- | --- | --- |
|  | $S_f=2$ | $S_f=100$ | $S_f=1000$ | $S_f=2$ | $S_f=100$ | $S_f=1000$ |
| **#Words unstemmed** | 19,588 | 1,379 | 85 | 16,872 | 511 | 8 |
| **#Words stemmed** | 13,519 | 1,179 | 119 | 12,856 | 571 | 15 |

Table 5. Number of frequent stemmed and unstemmed words from two corpora.

| | | | | | | | | | |
|---|---|---|---|---|---|---|---|---|---|
| studi | 5529 | aim | 1946 | product | 1500 | role | 1227 | retract | 1909 |
| chang | 5212 | area | 1901 | high | 1483 | howev | 1220 | use | 1783 |
| project | 4893 | organ | 1893 | region | 1469 | affect | 1198 | studi | 1773 |
| popul | 4225 | result | 1886 | interact | 1460 | method | 1198 | express | 1714 |
| environ | 4065 | experi | 1864 | forest | 1455 | factor | 1179 | effect | 1649 |
| differ | 3975 | impact | 1849 | approach | 1439 | recent | 1171 | protein | 1626 |
| ecosystem | 3335 | water | 1819 | two | 1429 | main | 1162 | group | 1607 |
| develop | 3305 | propo | 1818 | object | 1400 | particular | 1157 | signif | 1251 |
| model | 3193 | new | 1817 | condit | 1383 | current | 1156 | increa | 1234 |
| plant | 3184 | determin | 1801 | conserv | 1372 | activ | 1150 | result | 1170 |
| effect | 3132 | function | 1799 | analysi | 1366 | anim | 1128 | gene | 1148 |
| climat | 2780 | may | 1759 | predict | 1355 | larg | 1128 | human | 1010 |
| can | 2757 | habitat | 1716 | within | 1344 | knowledg | 1115 | | |
| genet | 2752 | increa | 1703 | potenti | 1316 | rang | 1112 | | |
| import | 2743 | gene | 1672 | base | 1311 | key | 1086 | | |
| research | 2692 | one | 1661 | adapt | 1307 | field | 1083 | | |
| biodiv | 2665 | assess | 1638 | year | 1307 | molecular | 1077 | | |
| also | 2574 | test | 1624 | identifi | 1303 | allow | 1076 | | |
| data | 2448 | level | 1622 | pattern | 1289 | mechan | 1069 | | |
| ecolog | 2415 | structur | 1611 | work | 1286 | anali | 1066 | | |
| manag | 2238 | marin | 1599 | dynam | 1280 | scale | 1065 | | |
| system | 2213 | select | 1598 | global | 1266 | present | 1049 | | |
| diver | 2177 | includ | 1584 | fish | 1260 | resourc | 1045 | | |
| process | 2157 | biolog | 1565 | mani | 1259 | control | 1035 | | |
| communiti | 2130 | group | 1564 | evolut | 1258 | human | 1034 | | |
| understand | 2123 | inform | 1560 | carbon | 1256 | evolutionari | 1030 | | |
| natur | 2107 | time | 1538 | well | 1246 | | | | |

Table 6. Item sets of very-frequent stemmed words with threshold $S_f=1000$ of documents from PM and BD corpora (only words having at least three characters are kept).

We can see (Table 5 and Table 6) that considering the "true" corpus the amount of frequent terms is a good signal for interpreting the biodiversity domain. Nevertheless, for the "false" corpus the amount is a small signal only indicating that the majority of documents talk about cell biology and medicine.

### 2.3 Nearest-neighbour analysis

Now we turn to a macroscopic analysis of corpora. Many clustering algorithms lead to a summarization of similarities between bags of words, one of them reveals close-in-context items within their collocations: k-nearest-neighbour algorithm (KNN). It was created by [Cover and Hart, 1967] and leads to good results with different kinds of data. We can argue that the frequent itemset extraction method of [Agrawal and Srikant, 1994] called apriori is a variant of KNN. An interesting property of this kind of algorithm is the low-level time-complexity. It is also efficient with sparse data like text data. To visualize a large global clustering we used the *Igraph* package implemented for large network analysis and visualization [Csardi and Nepusz, 2006]. Thirteen layout algorithms are available. We specifically used the Fruchterman-Reingold layout, which is force-based combining attractive forces of adjacent vertices, and repulsive forces on all vertices [Fruchterman and Reingold, 1991]. We also used the DrL layout (*Distributed Recursive Layout*), also force-based and using the VxOrd routine offering a multi-level recursive version to obtain a better layout on big graphs, and the ability to add new nodes to a graph already displayed [Martin et al, 2011].

*Definition 3: Data Structure*
Let $M = (m_{ij})_{1 \leq i \leq n, 1 \leq j \leq m}$ be a data matrix where *i* represents i-th line and so the word *i*; *j* represents j-th column, hence the document *j*; and *n* is the number of words and *m* the number of documents.

*Definition 4: Neighbourhood*
Two items $w_i$ and $w_k$ are neighbours if it exists in document $d_j$, where $w_i \in d_j$ and $w_k \in d_j$.

We previously discussed that very-frequent words are interesting to extract. We want now not only to look at a set of items but their relationships, and especially as a first step how this global set of relationships is featured. Visualization is a good tool to fill this task because thousands of relationships are involved and no primary criteria permit selection of a pool of specific or more relevant relationships. If we try to visualize the symmetric data matrix of most frequent terms between each other for instance, we get a ball of links without structural specificity; each items having the whole set of items as the nearest neighbours.
For improving clustering efficiency we need to operate a data reduction. The algorithm below shows a reduction by the weighted margin mean. Computing the incidency matrix is based on a simple reduction by subtracting means of non-null values of each line to the matrix value of the same line.

*Definition 5: Data Reduction*
$$Y = M.\ ^tM - \beta. <M>_l$$
where <*M*>*l* is a mean vector of a line from M. $\beta$ plays as a regulation factor to regulate the rate of nearest neighbours, in fact the number of nearest neighbours is not defined explicitly.

Nearest-Neighbour Algorithm
**Input**:
    M: a sparse matrix terms x documents, with dim(M)=(n,m) such that M[i,j] is the number of occurrences of a term i in the document j,
    Min: minimum frequency
    Max: maximum frequency
    Beta: scaling factor
    Binary: 0 if real data, 1 if data are binary
**Output**: Layout in 2-Dimensions
*# layout with Fruchterman algorithm*
1:     Create a vector V, with dim(V)=n such that RowSums(M[i,])<= Max and RowSums(M[i,]) >= Min
2:     M' = M[V > 0]
3:     Create is a matrix terms x terms: TD = M' * t(M')
4:     Compute Vm the mean vector by line with dim(Vm) = n such that Vm[i] = mean(M[i,]) with M[i,j] =/= 0 for all j
5:     **IF** Bin = 1 Make scaling operation, TD_norm = TD – Beta*Vm
              TD_ norm = TD_norm>=0
       **ELSE**     GoodVal = TD >= -1*Beta*Vm & TD <= +1*Beta*Vm
               TD_norm = TD * good_value
6:     Binary transform, TD_norm[TD_norm > 0] = 1
7:     Keep positive values TD = TD_norm[ rowSums(TD_norm) > 0]
8:     Compute the mean of links per node, Nb_mean_link = mean(rowSums(TD))
9:     Generate the layout Fruchterman for display with TD as adjacency matrix.
*# layout with DRL*
10:     Create a vector V, with dim(V)=n such that V[i] <= Max and V[i] >= Min

| | | |
|---|---|---|
| 11: | Create a binary clone M of M, M' = M[M > 1] <- 1 | |
| 12: | Create a vector V, with dim(V)=n such that RowSums(M'[i,]) <= Max and RowSums(M'[i,]) >= Min | |
| 13: | M'' = M[V > 0] | |
| 14: | Create matrix terms x terms: TD = M'' * t(M'') | |
| 15: | TD' = TD [TD > 1] <- 1 | |
| 16: | Compute Vm the mean vector by line with dim(Vm) = n such that Vm[i] = mean(TD'[i,]) with TD'[i,j] =/= 0 for all j | |
| 17: | **IF** Bin = 1 Make scaling operation, TD_norm = TD' – Beta*Vm | |
| | TD_pos = TD_norm>=0 | |
| | **ELSE** GoodVal = TD' >= -1*Beta*Vm & TD' <= +1*Beta*Vm | |
| | TD_norm = TD * good_value | |
| 18: | Binary transform, TD_norm[TD_norm > 0] = 1 | |
| 19: | Keep positive values TD = TD_norm[rowSums(TD_norm) > 0] | |
| 20: | Compute the mean of links per node, Nb_mean_link = mean(rowSums(TD)) | |
| 21: | Generate the layout DRL for display with TD as adjacency matrix. | |

We used the Fisher's Iris dataset to validate the clustering approach. The dataset consists of 150 individuals described by five features and forming three classes (Table 7). Focusing on two classes (*versicolor* and *verginica*) only one feature makes a fine-grained discriminant classification (*Petal.width*); for sure, the usage of a value mean with features is not able to capture this difference. Hence the algorithm presented above can only discriminate two classes as seen in Figure 2.

| | Sepal.Length | Sepal.Width | Petal.Length | Petal.Width |
|---|---|---|---|---|
| **Setosa** | 5.006 | 3.428 | 1.462 | 0.246 |
| **Versicolor** | 5.936 | 2.770 | 4.260 | 1.326 |
| **Verginica** | 6.588 | 2.974 | 5.552 | 2.026 |

Table 7. Mean values about Iris dataset for each class.

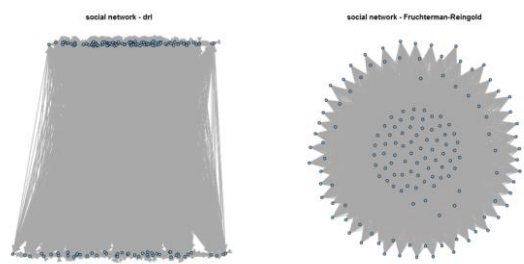

Figure 2. Display of Iris dataset: DRL (left) and Fruchterman-Reingold (right).

Our hypothesis, through visualization, aims at comparing different ranges of word frequency and at distinguishing their impact on global classification. Basically we could guess, on the one hand, that lexical items contribute equally each one to clustering. Even more we can suppose that more frequent words are more clustered than low frequent ones. On the other hand, we also could expect that "true" data (i.e. corpus BD) are quite more clustered than "false" data (i.e. corpus PM).

As the Zipf distribution shows (Figure 1) the range frequency can be considered as a good parameter to categorize numerically the lexical space. It is possible to define a partition of contiguous ranges depending upon the two first ranges and containing almost the same number of contexts.

*Definition 6: Context*
The context of a lexical item is a text area in which can be seen an occurrence of a lexical item. Let $w_1$ and $w_2$ be two lexical items. If $f_1$ and $f_2$ are, respectively, the frequency for each lexical items, $C = f_1 + f_2$ is the total number of contexts.

For instance three lexical items having frequency two generate six contexts. Regarding corpus BD, Table 8 shows that a series of frequency ranges from which the first two ones are [2-5] and [6-12] produce 23 ranges and have in average 33,554 contexts.

| 1st frequency range | 2 -- 2 | 2 -- 3 | 2 -- 5 | 2 -- 9 | 2 -- 20 |
|---|---|---|---|---|---|
| 2nd frequency range | 3 -- 3 | 4 -- 5 | 6 -- 11 | 10 -- 27 | 21 -- 60 |
| averaged #contexts | 6987 | 14547 | 26534 | 40911 | 58377 |
| #range | 42 | 20 | 11 | 6 | 5 |
|  |  |  |  |  |  |
| 1st frequency range | 2 -- 2 | 2 -- 3 | 2 -- 5 | 2 -- 9 | 2 -- 20 |
| 2nd frequency range | 3 -- 3 | 4 -- 6 | 6 -- 12 | 10 -- 25 | 21 -- 65 |
| averaged #contexts | 9233 | 19200 | 33554 | 55331 | 96860 |
| #range | 83 | 40 | 23 | 14 | 8 |

Table 8. Number of frequency ranges depending on the context size of the first two ones (upper table, Corpus PM; bottom table, corpus BD).

Let $K$ be a granularity factor (number of ranges) and $N_c$ the averaged number of contexts per range, we observe that:

$$K * N_c = Cst \pm 0.001\%$$

## 2.4 Frequency range analysis

Global visualization changes when we select a set of lexical items from different ranges. Figure 3 shows clustering drawings with range [2-5], Figure 4 with range [2-3], Figure 5 with range [2-20] and Figure 6 with range [2-2].

Choosing the range [2-9], the equipartition series gives eight ranges. Observing visualization for different ranges (Figures 3, 4, 5 and 6) we argue that the density of clusters evolve closely in regard to the size of the frequency range and number of ranges in associated contexts (Table 8). We also observe density of high-frequency words clustered together is different than low-frequency words clustered together. It seems that more frequent words reduce density of low-frequent ones in terms of class.

From studies about argumentation scheme linking lexical items such as verbs, connectors and nouns or adjectives from the corpus BD we get a list of useful verbs for technical argumentation in scientific discourse. This set consists of 291 verbs and 705 different tokens (gerund, past…). In the figures, points associated to one of the verb lists are coloured in red. Several tens of verbal forms belong to the clustered area as well as for corpus PM and corpus BD. It means that some verbs are deliberately useful to argumentation to this or that technical context. But some red points can be seen near dense areas. It means clearly a polysemy of verbs playing a role in different contexts.

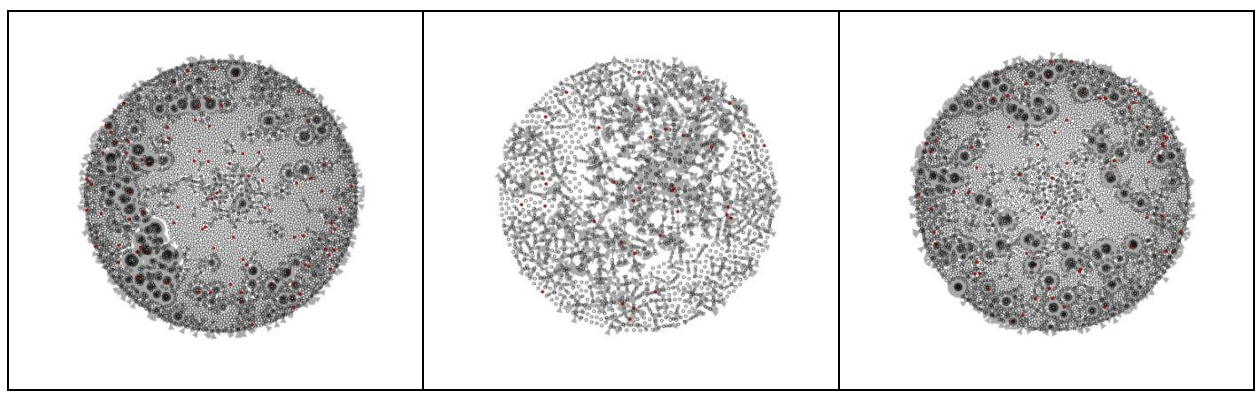

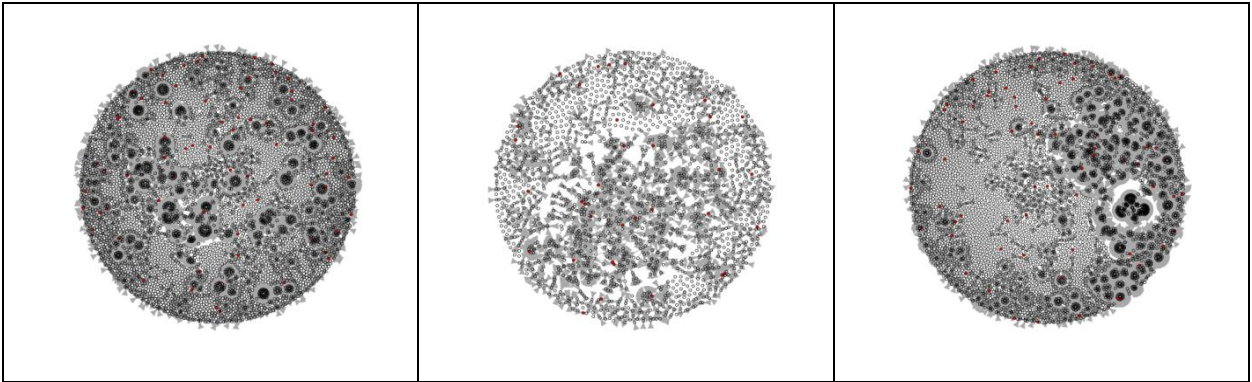

Figure 3. At top line, display of unstemmed words from the corpus PM with Fruchterman layout: frequency range [2-9], β=3, <neighbours>=0, N=5375 words (top left) frequency [6-9], β=2, < neighbours >=2, N=1739 words (top middle) frequency [2-5], β=2, < neighbours >=1, N=4751 words (top right). At bottom line, display of unstemmed words from the corpus BD with Fruchterman layout: frequency [2-9], β=3, < neighbours >=1, N=6097 words (bottom left) frequency [6-9], β=2, < neighbours >=1, N=2022 words (bottom middle) frequency [2-5], β=2, < neighbours >=1, N=5395 words (bottom right).

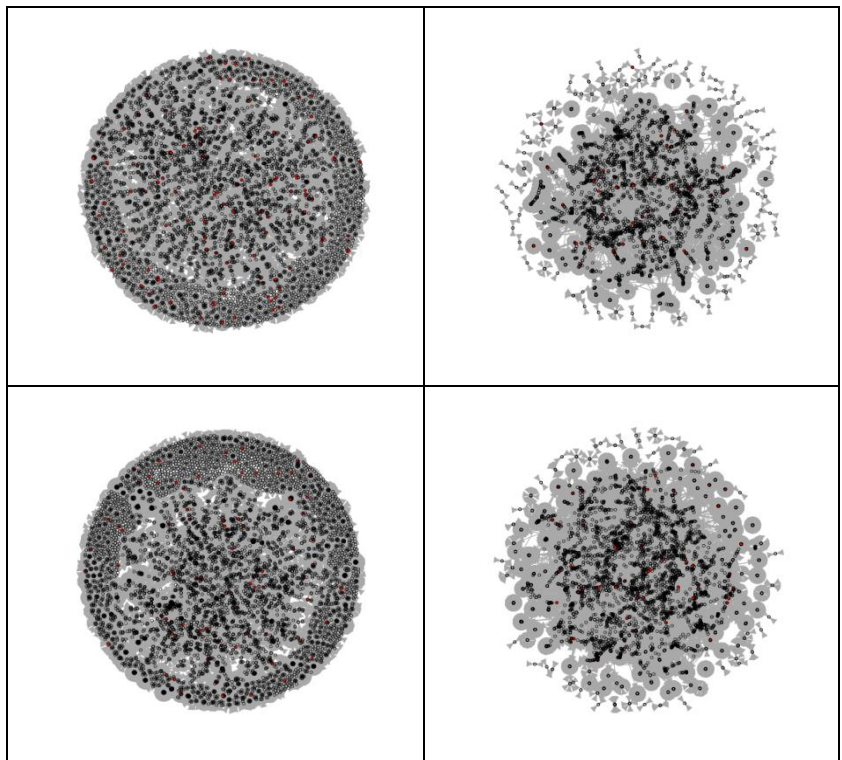

Figure 4. At top line, display of unstemmed words from the corpus PM: frequency [2-3], β=1, < neighbours >=7, N=4245 words (DRL, top right), β=1, < neighbours >=3, N=5708 words (Fruchterman, top left).
At bottom line, display of unstemmed words from the corpus BD: frequency [2-3], B=1, < neighbours >=8, N=5698 words (DRL, bottom right), β=1, < neighbours >=4, N=6938 words (Fruchterman, bottom left).

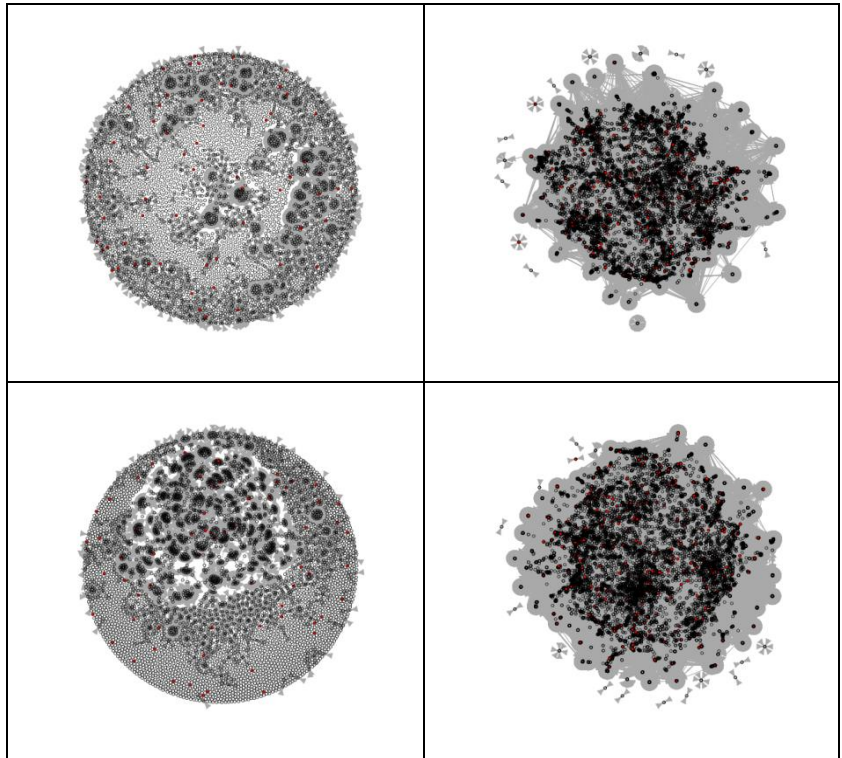

Figure 5. At top line, display of unstemmed words from the corpus PM: frequency [2-20], β=1, < neighbours >=45, N=8278 words (DRL, top right), β=5, < neighbours >=1, N=4479 words (Fruchterman, top left). At bottom line, display of unstemmed words from the corpus BD: frequency [2-20], β=1, < neighbours >=31, N=13997 words (DRL, bottom right), β=5, < neighbours >=1, N=5841 words (Fruchterman, bottom left).

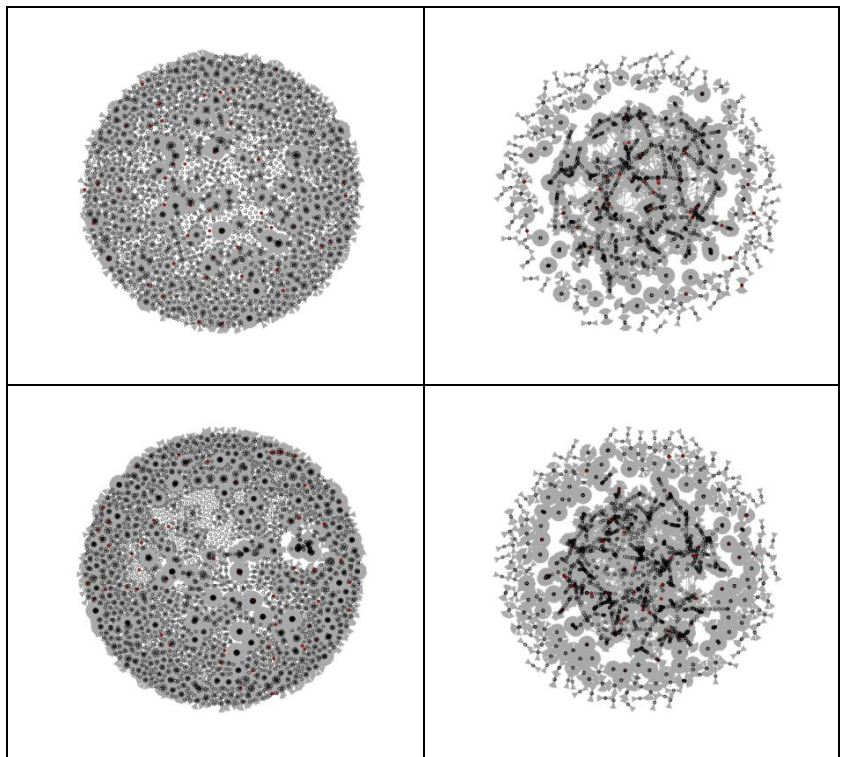

Figure 6. At top line, display of unstemmed words from the corpus PM: frequency [2-2], β=1, < neighbours >=4, N=2328 words (DRL, top right), β=1, < neighbours >=2, N=2962 words (Fruchterman, top left). At bottom line, display of unstemmed words from the corpus BD: frequency [2-2], β=1, < neighbours >=5, N=3437 words (DRL, bottom right), β=1, < neighbours >=3, N=3871 words (Fruchterman, bottom left).

## 2.5 Corpora comparison

If we look at the global structure for a given corpus and a large range of frequencies [2-20] we should be able to compare corpora to each other. Hence both nearest neighbourhood display (NND) and Zipf distribution appear to be a measure of comparison for different kinds of corpus.

Figures 7, 8 and 9 show this measure for biodiversity corpus (BD), patents corpus (TC) and scientific articles corpus (SCI). This set of corpora is representative of the technical discourse. Zipf distributions are equivalent but NND is a little bit different though very similar. Texture is high-grained granular for a third of the surface and low-grained granular for another third part–the rest is distributed widespread. For the TC corpus, granularity is thinner showing a less shared discourse. BD and SCI discourses are more shared and well clustered, perhaps as well as research communities.

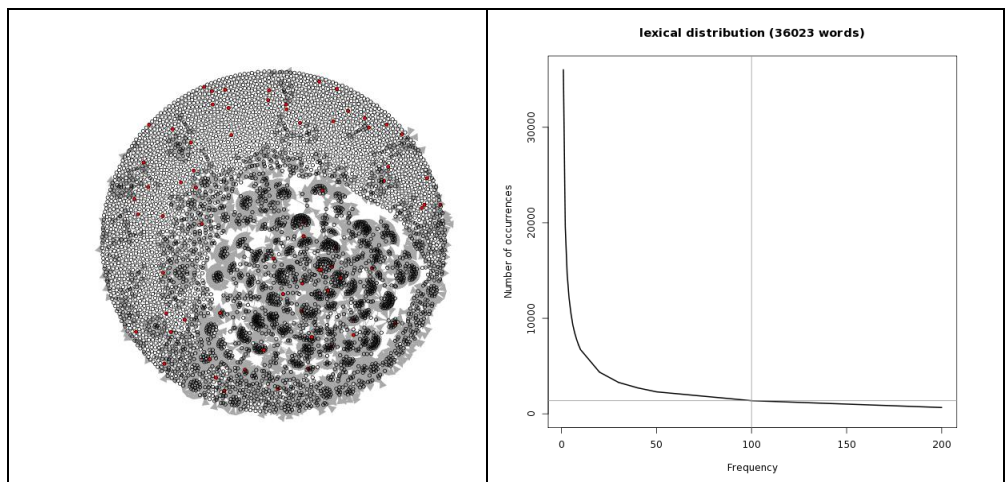

Figure 7. BD corpus, frequency [2-20]; β=5, < neighbours >=1, N=4985 words.

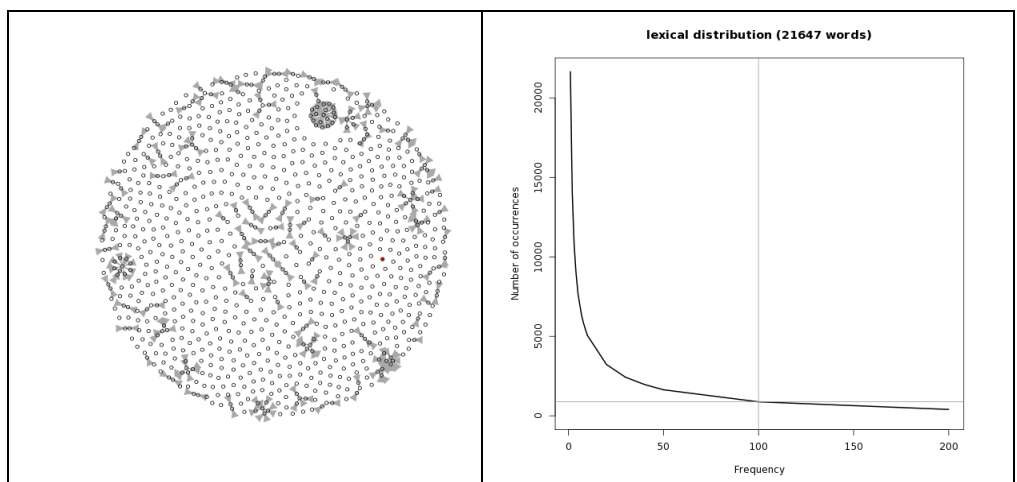

Figure 8. TC corpus, frequency [2-20]; β=1, < neighbours >=8, N=10434 words.

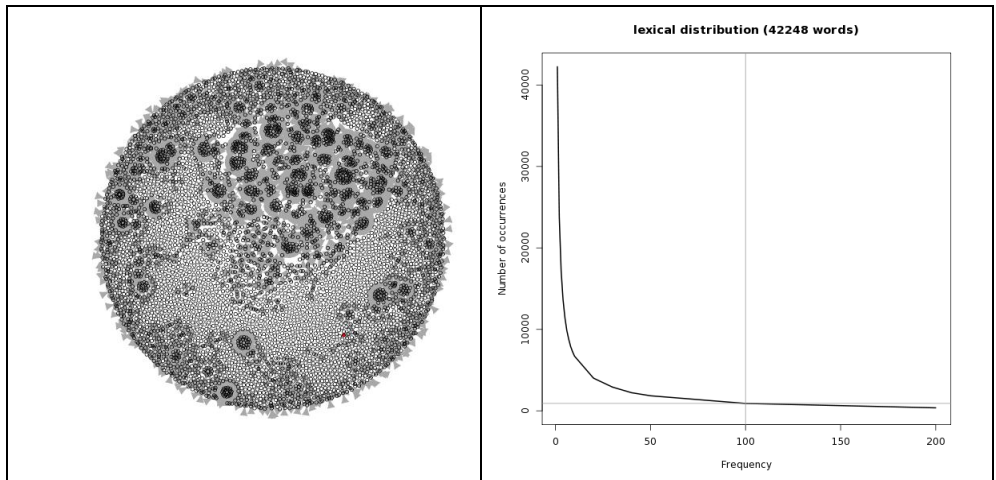

Figure 9. SCI corpus, frequency [2-20]; β=5, < neighbours >=1, N=5225 words.

The following corpus is built from Reuter's news. This type of corpus is typical of daily news and therefore a general language vocabulary. We can see in Figure 10 that the structure of NND and Zipf distribution are not far from those of technical discourse.

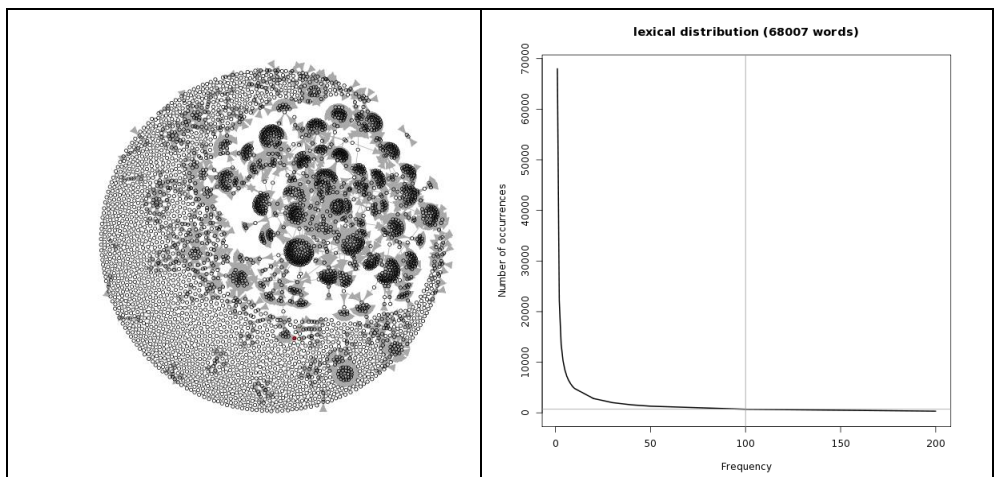

Figure 10. RT corpus, frequency [2-20]; β=5, < neighbours >=1, N=3946 words.

The third group of corpora gathers the Twitter corpus and the newsgroup corpus. These two corpora can be seen as we find short speech (in the same way there is chat and sms communication). The Zipf distribution is similar than the technical and General Language groups but NND is not really clustered.

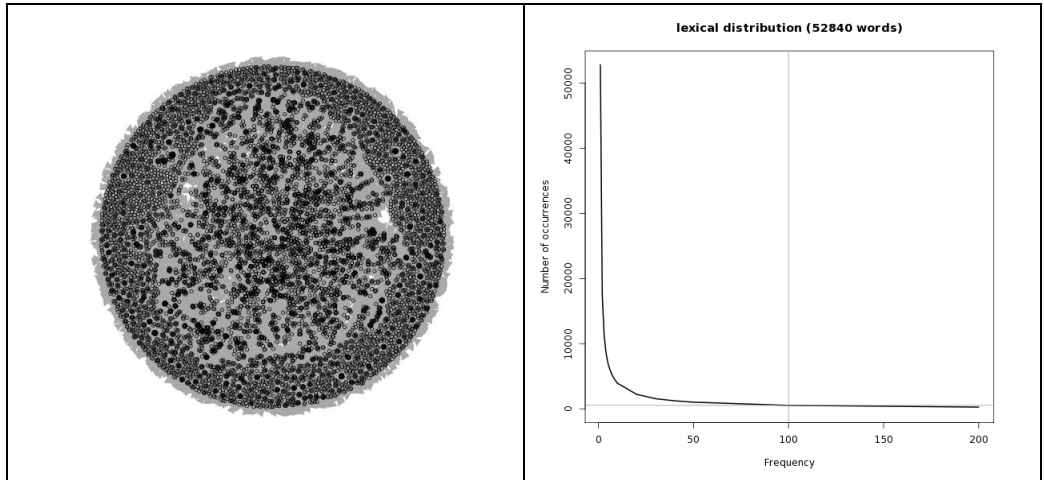

Figure 11. TW corpus, frequency [2-20]; β=1, < neighbours >=3, N=8908.

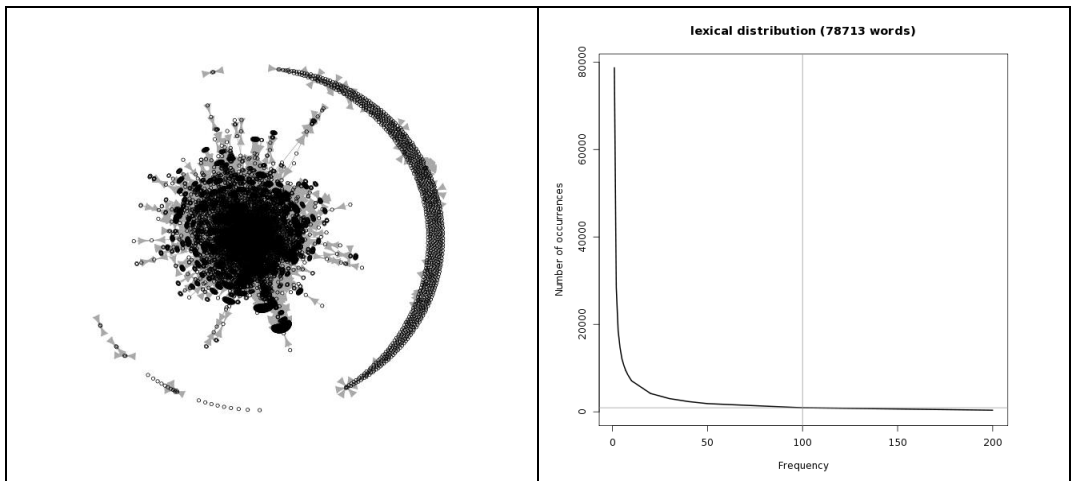

Figure 12. NG corpus, frequency [2-20]; β=5, < neighbours >=5, N=12930.

Figures 13, 14, 15, 16 and 17 show the results of the fourth (and last) group of corpora. Corpora RD, SL, BL and CS are generated documents. In Figure 17 we see NND regarding rejected documents from PubMed (PM corpus). In the PM corpus documents have been written as if they were real scientific facts but the arguments (or one argument) cannot validate the status as scientific articles. Thus syntax and semantic are valid but not the pragmatic level (reasoning). We see that NND and Zipf distribution are strangely very close to those already previously seen for technical corpora. It means that reasoning and community-based selection (pragmatic level) of an article plays a central role in making its status. In regard to other corpora the Zipf distribution is clearly degenerated. NND are different than others. The CS corpus is more troubling and can be more or less similar to the TC corpus with a sparse clustering texture.

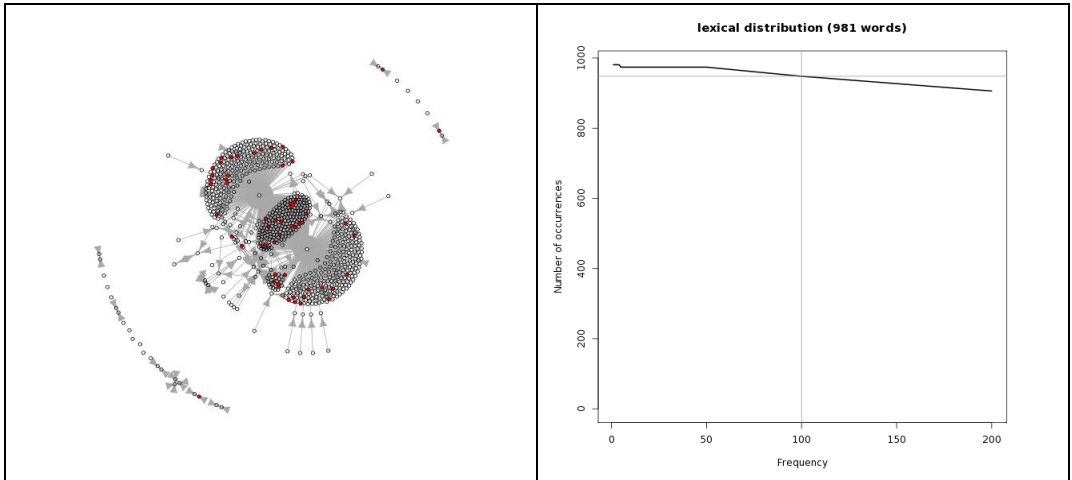

Figure 13. RD corpus, frequency [2-20]; β=7, < neighbours >=1, N=774.

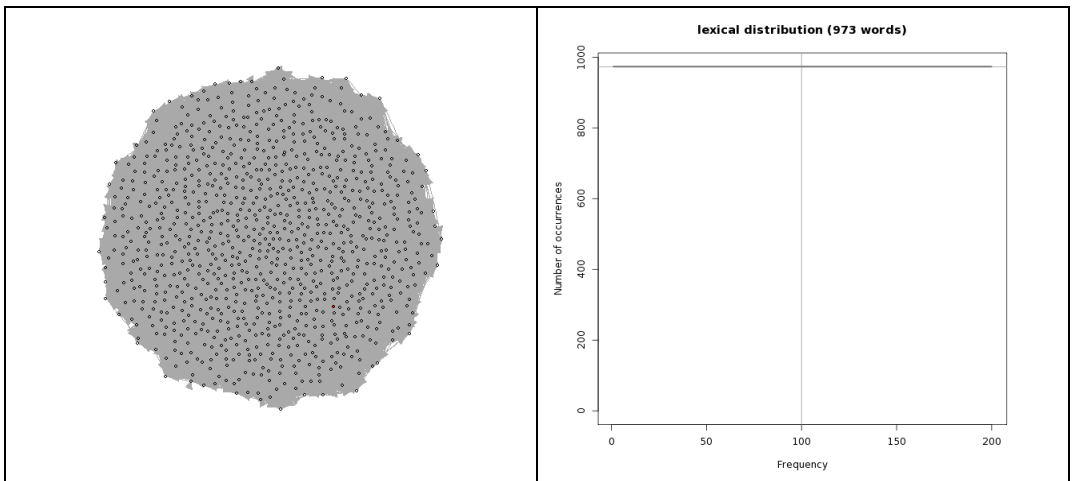

Figure 14. SL corpus, frequency [2-20]; β=1, < neighbours >=391, N=972.

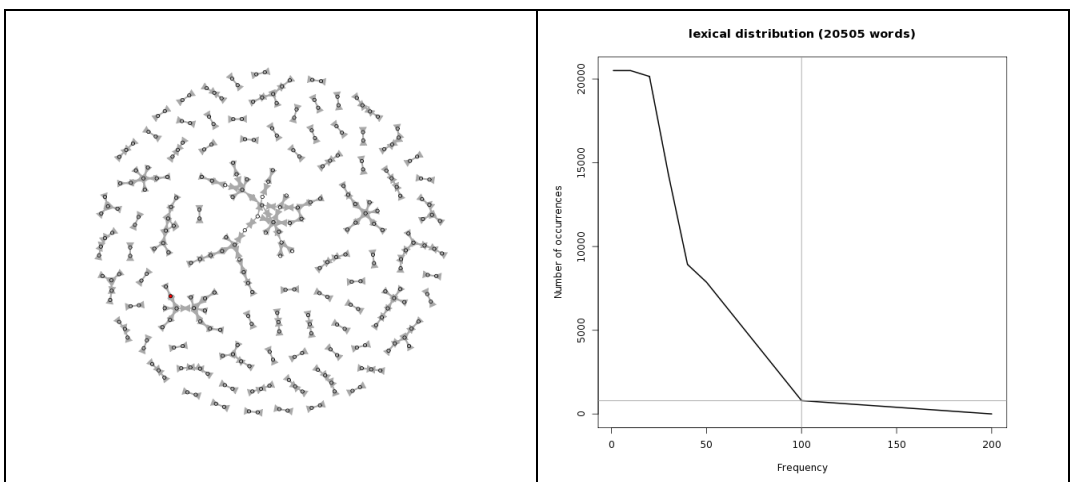

Figure 15. BL corpus, frequency [2-20]; β=1, < neighbours >=1, N=282.

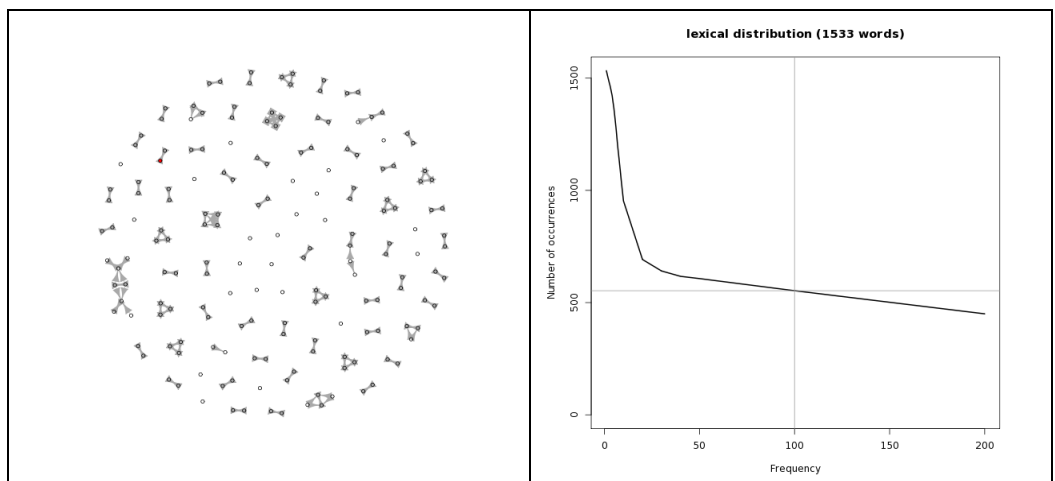
Figure 16. CS corpus, frequency [2-20]; β=1, < neighbours >=1, N=174.

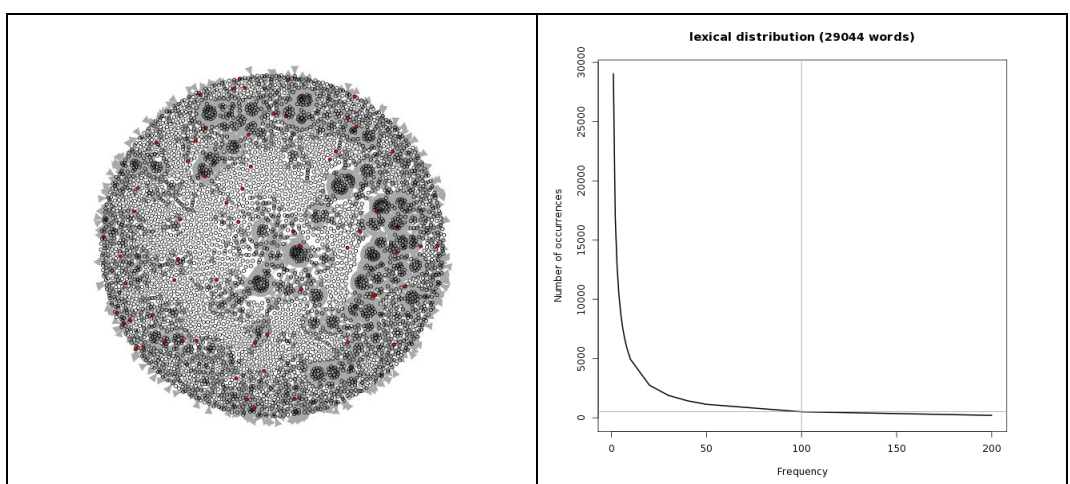
Figure 17. PM corpus, frequency [2-20]; β=5, < neighbours >=1, N=4479.

## 2.6 Discussion

Some sciences try to learn close associations between components. We can cite chemistry and sociology. Some other sciences try to learn more about global structures like economics and astrophysics. Computational linguistics and lexical statistics are domains able to take an overview of a whole set of relationships, as well as focus on specific relationships. In this chapter we try to show some results for a whole overview of closed relationships in similar short documents, highlighted by some items involved in specific argumentative relationships. Firstly, we try to explain how contradictions can occur explicitly in texts as specific relationships. Secondly, two corpora have been studied to extract global information. They share common properties such as: short document size, technical domain, English language, natural distribution of lexical items, and corpus size. Nevertheless, one corpus is a domain studied by many people as a scientific active domain (i.e. biodiversity), the other consists of "hoax" documents written by people as true documents. Surprisingly, there is striking similarity of global clustering visualization between hoax documents and true documents. The texts are natural language factual information interpreted by humans. Originally experimental data generated or pretreated prior to lead to published interpretations. In the next chapter we try to highlight interpretation lacks of evidence in several areas. Beyond syntactic associations humans choose their words based on their understanding that cannot be stable and give rise to divergent views even paradox or contradiction.

**Conclusion**
In this paper we present two sources of contradictions occurring in text data. The first one is purely syntactic related to the writer's intention in his or her article. The second one is related to factual data, obtained from experiments and leading to an elementary basis of interpretation. The goal of the paper influences how an article is written. When intention of the writer is ethically related to the scientific concern of truth, ambiguity relied only on difficulties to obtain unambiguous fact data from experiments entached by noise or lack of up-to-date categories to help interpretation. When intention of a writer is motivated by his or her career and reputation improvement, data is not central in playing a role but only rhetorical discourse of the writer, leading to improper relations but explained in the same way as real facts.

We used a set of papers in biology to identify the source of uncertainty arising in interpretation by humans of relationships in data. Secondly, ambiguities in natural language caused by its inherent polysemy are discussed through a nearest-neighbour display associated to a Zipfian distribution to compare structural contents of corpora. Four kinds of discourse (technical, general, short-communication and artificial) have been studied. Technical discourse has a similar structure than general discourse; short-communication and artificial are often dissimilar both with neighbourhood and frequent used distribution. But one artificial corpus with invalid technical facts lead to a similar structure than technical and general discourse; meaning that pragmatic and social control regarding reasoning makes sense for technical documents potentially containing contradictory arguments.

In this study we tried to point out the source of uncertainty to interpret relationships in data. Traditionally, dictionaries are used to disambiguate or state ambiguity. Syntax is pointed out, or even semantics. In this paper we claim that pragmatics also play a big role, and even a priori knowledge is not sufficient because of socio-semantic consensus about concepts. We did not propose a controlled process to subtract noise from data; leaving out bad intention of a writer or uncertainty of data leading to contradictory interpretation. It should be a serious issue to make a corpus cleaner for concept and name entity extraction and their relationships.